# Convergence of the EM Algorithm for Gaussian Mixtures with Unbalanced Mixing Coefficients

Iftekhar Naim                                                                  INAIM@CS.ROCHESTER.EDU
Daniel Gildea                                                                  GILDEA@CS.ROCHESTER.EDU
Department of Computer Science, University of Rochester
Rochester, NY 14627, USA

## Abstract

The speed of convergence of the Expectation Maximization (EM) algorithm for Gaussian mixture model fitting is known to be dependent on the amount of overlap among the mixture components. In this paper, we study the impact of mixing coefficients on the convergence of EM. We show that when the mixture components exhibit some overlap, the convergence of EM becomes slower as the dynamic range among the mixing coefficients increases. We propose a deterministic anti-annealing algorithm, that significantly improves the speed of convergence of EM for such mixtures with unbalanced mixing coefficients. The proposed algorithm is compared against other standard optimization techniques like BFGS, Conjugate Gradient, and the traditional EM algorithm. Finally, we propose a similar deterministic anti-annealing based algorithm for the Dirichlet process mixture model and demonstrate its advantages over the conventional variational Bayesian approach.

## 1. Introduction

Clustering is a widely used exploratory data analysis tool that has been successfully applied to biology, social science, information retrieval, signal processing, and many other fields (Jain et al., 1999). In many of these applications (for example, biological data analysis, anomaly detection, image segmentation, etc.), the goal is to identify rare groups or small clusters (in terms of number of members) in the presence of other larger clusters. In this paper, we focus on the particular problem of clustering large datasets with high dynamic range in cluster sizes.

The Gaussian mixture model is a powerful model for data clustering (McLachlan & Peel, 2000). It models the data as a mixture of multiple Gaussian distributions where each Gaussian component corresponds to one cluster. Let $\mathbf{X} = \{\mathbf{x}_1, \mathbf{x}_2, \ldots, \mathbf{x}_N\}$ be $N$ i.i.d. random vectors that follow a $K$ component Gaussian mixture distribution. The $j^{th}$ Gaussian component in the mixture is defined by its mean $\boldsymbol{\mu}_j$, covariance $\boldsymbol{\Sigma}_j$, and the mixing coefficient $\alpha_j$ ($\alpha_j > 0$ and $\sum_{j=1}^{K} \alpha_j = 1$). These parameters together are represented as the parameter vector $\Theta = [\alpha_j, \boldsymbol{\mu}_j, \boldsymbol{\Sigma}_j]_{j=1}^{K}$. Our goal is to estimate model parameters $\Theta$ given the data $\mathbf{X}$. The parameter estimation is typically accomplished by the Expectation-Maximization (EM) algorithm (Dempster et al., 1977). In this paper, we empirically show that EM exhibits slow convergence if one of the Gaussian mixture components has a very small mixing coefficient compared to others, and there exists some overlap among the mixture components. We explain this slow convergence of EM for a mixture with unbalanced mixing coefficients using the convergence analysis framework presented by Xu and Jordan (1996), and Ma et al. (2000).

We present a solution to this problem based on Deterministic Annealing (Rose, 1998; Ueda & Nakano, 1998). It is well known that deterministic annealing can help prevent the EM algorithm from getting trapped in local optima (Ueda & Nakano, 1998). Traditional deterministic annealing follows a monotonically decreasing temperature schedule by slowly increasing the inverse temperature parameter ($\beta$) from 0 to 1. We propose a novel non-monotonic temperature schedule that can improve the speed of convergence as well. We call this modified annealing schedule *Anti-annealing*. We start with a traditional temperature

---

schedule by slowly increasing $\beta$ from 0 to 1. Next we continue increasing $\beta$ beyond 1, up to a chosen upper bound, and finally slowly decrease $\beta$ down to 1. Our experiments demonstrate the effectiveness of the proposed Anti-annealing schedule to improve the speed of convergence of the EM algorithm for unbalanced mixtures. Finally, we extend our results to the Dirichlet Process Gaussian Mixture Models (DP-GMM).

## 2. Convergence of EM Algorithm

### 2.1. Related Work

The EM algorithm is guaranteed to monotonically converge to local optima under mild continuity conditions (Dempster et al., 1977; Wu, 1983). Redner and Walker (1984) show that EM has linear rate of convergence and suggest that Newton's or Quasi-Newton methods should be preferred over the EM algorithm. They also experimentally show that EM converges slowly in the presence of overlapping clusters. Xu and Jordan (1996) analyze the rate of convergence of the EM algorithm and show that EM exhibits a super-linear rate of convergence as the overlap among the mixture components goes to zero (Xu & Jordan, 1996). They forge a connection between the EM algorithm and gradient ascent and prove that rate of convergence of the EM algorithm depends on the condition number of a projected Hessian matrix $E^T P(\Theta^*) H(\Theta^*) E$, where $\Theta^*$ is the optimum parameter value, $E = [e_1, \ldots, e_m]$ is a set of unit basis vectors spanning the constrained parameter space (satisfying the constraint $\sum_{j=1}^K \alpha_j = 1$), $P(\Theta^*)$ is a projection matrix, and $H(\Theta^*)$ is the Hessian of the log-likelihood function. Later, Ma et al. (2000) extend this result and show that the rate of convergence of EM is a higher order infinitesimal of maximum pairwise overlap among the mixture components. Salakhutdinov et al. (2003) propose the Expectation Conjugate Gradient (ECG) algorithm for highly overlapping Gaussian mixtures. ECG allows faster convergence than the traditional EM algorithm in the presence of high overlap among mixture components. Therefore, the impact of component overlap on the convergence of EM has been known for more than a decade. While overlap among components is the major factor that influences rate of convergence of EM, we show that mixing coefficients can also significantly influence the speed of convergence in the presence of some overlap. To our knowledge, this is the first work that addresses the impact of mixing coefficients on the rate of convergence of EM.

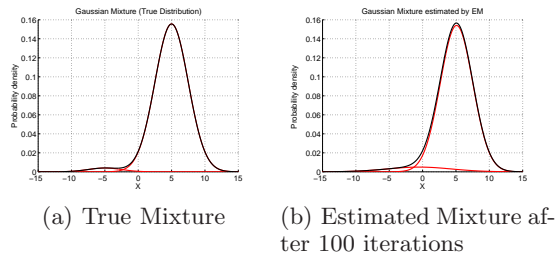

(a) True Mixture  (b) Estimated Mixture after 100 iterations

Figure 1. Performance of EM for two component mixture of Gaussians. The small cluster parameters did not converge to true values.

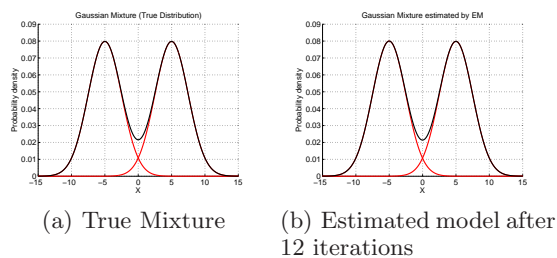

(a) True Mixture  (b) Estimated model after 12 iterations

Figure 2. EM converges quickly to correct parameters only after 12 iterations.

### 2.2. Convergence of EM for Small Clusters

We investigate the convergence of the EM algorithm for Gaussian mixtures with some small mixing coefficients. We start with 3 simulation examples. First, we create synthetic data with a 2-component univariate Gaussian mixture with the parameters $\alpha = (0.025, 0.975)^T$, $(\mu_1, \mu_2) = (-5.0, 5.0)$, $(\Sigma_1, \Sigma_2) = (6.25, 6.25)$ (as shown in Figure 1(a)). We perform 2-component EM fitting and after 100 iterations, we get a parameter estimate of $\hat{\alpha} = (0.059, 0.9410)^T$, $(\hat{\mu_1}, \hat{\mu_2}) = (-0.497, 5.08)$, $(\hat{\Sigma_1}, \hat{\Sigma_2}) = (23.15, 5.92)$. In this simple setting, the inaccurate result is due to the slow speed of convergence. If we allow EM to run for many more iterations, it usually converges to the true parameter values.

Next, we create another dataset (Figure 2(a)) with balanced mixing coefficients ($\alpha = (0.5, 0.5)^T$) and leave the other parameters the same. Under the same overlap (since means are not changed), EM converges to almost accurate parameters much faster, after only 12 iterations. This simulation implies that under the same amount of overlap, more balanced mixing coefficients yield faster convergence of EM. Finally, we create a synthetic dataset with skewed mixing coefficients, but relatively less overlap (Figure 3(a)). Means are set at points $(\mu_1, \mu_2) = (-10.0, 10.0)$, mixing coefficients at $\alpha = (0.025, 0.0975)^T$, and covariance values at $(\Sigma_1, \Sigma_2) = (6.25, 6.25)$. Despite having very small mixing coefficient values, EM converged to the true parameters in only 23 iterations. This example emphasizes the critical impact of overlaps on speed of convergence. However, the first example also implies that under some overlap, mixing coefficients have strong influence on the speed of convergence.

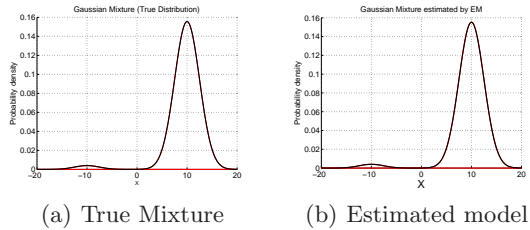

(a) True Mixture  (b) Estimated model

Figure 3. EM converges to correct parameters only after 23 iterations despite the highly unbalanced mixing coefficients.

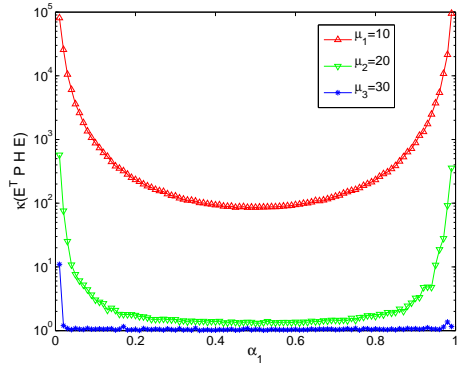

Figure 4. A plot showing the condition number of the effective Hessian matrix for the EM algorithm with varying mixing coefficient values. The condition number increases significantly for extreme values of $\alpha_1$ in case of the overlapping clusters.

### 2.3. Why Slow Convergence?

Next, we explore the reason that EM shows slow convergence in the presence of small clusters. We explain it using the framework proposed by Xu and Jordan (1996). Xu and Jordan (and later Ma et al. (2000)) have shown that the rate of convergence of EM is upper bounded by the norm $\|I + E^T P(\Theta^*) H(\Theta^*) E\|$ i.e. the condition number of the projected Hessian $\kappa \left[ E^T P(\Theta^*) H(\Theta^*) E \right]$. As the condition number $\kappa \simeq 1$, the log-likelihood surface is spherical and allows fast convergence. On the other hand, a larger value of the condition number of the effective Hessian implies an elongated log-likelihood function which can cause slow convergence for any linearly convergent method. Our simulation results show that, in the presence of some overlap, the condition number of the Hessian matrix increases as one of the mixing coefficients decreases. We verify this by computing the condition number of the projected Hessian matrix $E^T P(\Theta^*) H(\Theta^*) E$ at the true parameter values for 3 different 2-component mixture configurations. For all the mixture configurations, we keep one of the means fixed at $\mu_2 = 0.0$ and then we vary the mean of the other components to $\mu_1 = 10$, 20, and 30. The variance of the components are set to 9. Then, we vary the mixing coefficients of the first component $\alpha_1$ from 0.01 to 0.99 and compute the condition numbers for each case and plot them in Figure 4. The results are quite intuitive. In the case of non-overlapping clusters ($\mu_1 = 20$ or $\mu_1 = 30$), the condition number did not change much for extreme values of $\alpha_1$. However, for the overlapping case ($\mu_1 = 10.0$), the condition number of the projected Hessian became a lot larger for extreme values (both higher and smaller values) of $\alpha_1$. To confirm this intuition, we performed another simulation where we generated 2 different 2-component Gaussian mixtures with varying mixing coefficients and looked at their log-likelihood surface. For simplicity, we plot log-likelihood as a function of only the means of the mixture components, and assume the mixing coefficients and covariances to be fixed. Figure 5 shows the log-likelihood surfaces (on the right) for the corresponding Gaussian mixtures (on the left). As one of the mixing coefficients becomes significantly smaller compared to the others, the log-likelihood surface tends to become flatter and elongated, which is in agreement with our intuition.

## 3. Proposed Solution

In this section, we propose a variation of the Deterministic Annealing EM (Ueda & Nakano, 1998) to address the speed of convergence issue explained above. We compare the result with two other well-known optimization techniques: Expectation Conjugate Gradient (ECG) (Salakhutdinov et al., 2003) and Quasi-Newton Method (BFGS) (Liu & Nocedal, 1989).

### 3.1. Deterministic Anti-Annealing Expectation-Maximization

Deterministic Annealing is a well-known technique to improve the convergence behavior for non-convex optimization problems. Ueda and Nakano (1998) proposed a Deterministic Annealing Expectation Maximization (DAEM) algorithm that varies the temperature from high to low temperature, and deterministically optimizes the objective function at each temperature. Conventionally, DAEM is used to provide reliable global convergence. In EM, we estimate the

Convergence of the EM Algorithm for Gaussian Mixtures with Unbalanced Mixing Coefficients

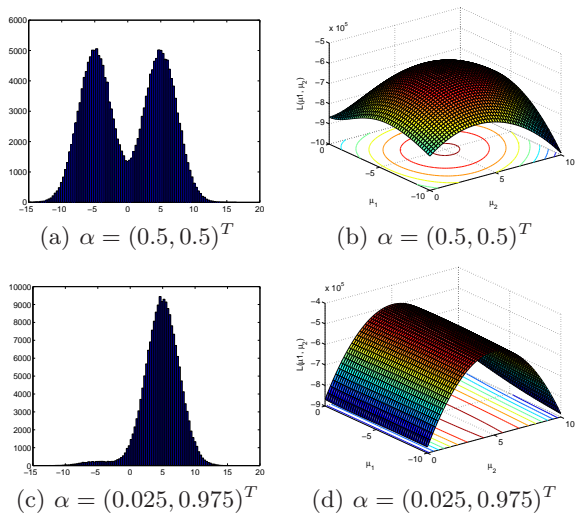

(a) $\alpha = (0.5, 0.5)^T$
(b) $\alpha = (0.5, 0.5)^T$
(c) $\alpha = (0.025, 0.975)^T$
(d) $\alpha = (0.025, 0.975)^T$

Figure 5. The geometry of the log-likelihood surface for 2-component Gaussian mixtures with different mixing coefficients, under the same amount of overlap.

posterior membership probabilities $h_j(t)$ (the probability that $\mathbf{x}_t$ belongs to the $j^{th}$ Gaussian component) using the following equation:

$$h_j(t) = \frac{\alpha_j P(\mathbf{x}_t|\boldsymbol{\mu}_j, \boldsymbol{\Sigma}_j)}{\sum_{i=1}^{K} \alpha_i P(\mathbf{x}_t|\boldsymbol{\mu}_i, \boldsymbol{\Sigma}_i)} \quad (1)$$

The DAEM algorithm modifies the E-step update rule using the scheduling parameter $\beta$:

$$h_j(t) = \frac{\left(\alpha_j P(\mathbf{x}_t|\boldsymbol{\mu}_j, \boldsymbol{\Sigma}_j)\right)^\beta}{\sum_{i=1}^{K} \left(\alpha_i P(\mathbf{x}_t|\boldsymbol{\mu}_i, \boldsymbol{\Sigma}_i)\right)^\beta} \quad (2)$$

In the M-step, the model parameters are estimated using this $h_j(t)$ values in exactly the same manner as the traditional EM algorithm. The parameter $\beta$ can roughly be interpreted as the inverse of temperature. DAEM typically starts at $\beta \simeq 0$ and slowly increases $\beta$ up to 1. At each $\beta$ value, DAEM iterates the E-step (2) and the M-step until convergence. The algorithm can be described as follows:

1. **Initialize:**
   - Set $\beta \leftarrow \beta_{min}$ ($0 < \beta_{min} \ll 1$)
   - Start with a random initial parameter $\Theta^{(0)}$

2. **Iterate until convergence:**
   - E-step: estimate posterior probabilities $h_j(t)$ using Equation 2.
   - M-step: estimate $\Theta^{(new)}$ using $h_j(t)$ values estimated in E-step

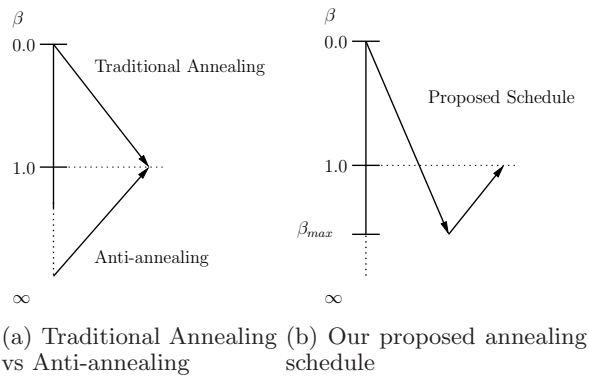

(a) Traditional Annealing vs Anti-annealing
(b) Our proposed annealing schedule

Figure 6. Scheduling of the DAEM algorithm. We chose to follow the one shown in (b) which allows both robust convergence (for $\beta < 1$) and faster convergence for $\beta > 1$.

3. Increase $\beta$

4. If $\beta \leq 1$, go back to step 2.

5. If $\beta > 1$, return.

Initially, for $\beta \rightarrow 0$, all the clusters are overlapping and the posterior probability becomes uniform, i.e. $h_j(t) \sim 1/K$ for all $j$. On the other hand, as $\beta$ becomes larger, DAEM allows less and less overlap and for $\beta \rightarrow \infty$ it becomes analogous to the winner-take-all (Kearns et al., 1998) approach, i.e. $h_j(t) = 1$ for only one $j$, and $h_i(t) = 0$, $\forall i \neq j$. For $\beta = 1$, it is reduced to the original posterior distribution given current $\Theta$.

Traditionally, DAEM starts with $\beta \simeq 0$ and slowly monotonically increases to $\beta = 1$. For tiny clusters, we observe that EM gives highly overlapping parameter estimates even after hundreds of iterations, and converges very slowly. We can improve the speed by starting from $\beta > 1$ and slowly decreasing it down to 1.0, where the objective function is same as that for EM. Let us call this scheduling strategy Anti-annealing scheduling. Although this improves the speed of convergence by restricting the amount of overlaps, the objective function becomes more irregular and EM tends to converge to poor local optima more frequently. In order to solve this problem, we follow a hybrid schedule, where we start with $\beta_{min} < 1$ and slowly increase it to a value $\beta_{max} > 1$ and then again decrease back to $\beta = 1$.

The proposed anti-annealing schedule (Figure 6(b)) improves the speed of convergence and helps avoid poor local optima as well. However, it is necessary to choose a slow enough temperature schedule, particularly for complicated data with a large number of clusters. We also need to perturb the estimated parameters after each iteration. At lower $\beta$ values, the effective number of clusters is often less than $K$, as many components share the same parameters. As we increase $\beta$, the clusters start to move around and effective number of clusters increases as a result of splitting. We must perturb the estimated parameters with a small noise term to enable the clusters to split. In our implementation, we add a small amount of random noise to the mean of each of the Gaussian components along the first principal component dimension of the associated cluster.

How Does Anti-Annealing Speed up Convergence?

Using the results shown by Ma et al. (2000), we explain the faster convergence of the proposed Anti-annealing method. Let $e_{ij}(\Theta^*)$ be the measure of overlap between the $i^{th}$ and $j^{th}$ Gaussian component:

$$e_{ij}(\Theta^*) = \begin{cases} \lim_{N \to \infty} \frac{1}{N} \sum_{t=1}^{N} h_i(t) h_j(t), \\ \quad \text{if } i \neq j \\ \\ \lim_{N \to \infty} \frac{1}{N} \sum_{t=1}^{N} (1 - h_i(t)) h_i(t), \\ \quad \text{if } i = j \end{cases}$$

and let the maximum component-wise overlap be: $e(\Theta^*) = \max_{i,j} e_{ij}(\Theta^*)$. Ma et al. (2000) show that the rate of convergence $r$ is a higher order infinitesimal of $e(\Theta^*)$:

$$r \leq \lim_{N \to \infty} \|I + E^T P(\Theta^*) H(\Theta^*) E\| = o(e^{0.5-\epsilon}(\Theta^*))$$

During the Anti-annealing phase ($\beta > 1$), the pairwise overlap values $e_{ij}(\Theta^*)$ decreases due to relatively harder membership probabilities. For the extreme case $\beta \to \infty$, the posterior probabilities $h_i(t)$ values are either zero or one, which causes $e_{ij}(\Theta^*) \simeq 0$ for all $i, j \in \{1, \ldots, K\}$ leading to superlinear convergence. In general, it can be easily shown that $e(\Theta^*)$ decreases as $\beta$ increases.

### 3.2. Expectation Conjugate Gradient (ECG)

The Conjugate Gradient method is known to outperform gradient descent methods for special cases when the objective function is elongated, and the conjugate direction is a better direction than the steepest gradient direction. Salakhutdinov et al. (2003) proposed an expectation conjugate gradient (ECG) method for optimizing the log-likelihood functions in the case of highly overlapping clusters. We evaluate the performance of the ECG algorithm for the small cluster scenario. The EM algorithm automatically satisfies several constraints on the parameter space: $\alpha_i \geq 0$, $\sum_i \alpha_i = 1$, and $\Sigma \succeq 0$. To ensure that the ECG algorithm satisfies the same set of constraints, Salakhutdinov et al. (2003) propose to re-parameterize the model parameters: $\alpha_j = \frac{e^{\lambda_j}}{\sum_l e^{\lambda_l}}$ and $\Sigma_j = L_j L_j^*$, where $L_j$ is the upper triangular matrix obtained by Cholesky decomposition of $\Sigma_j$. Due to space limitation, exact formulation of gradients under the re-parameterization is presented in the Appendix. We apply the standard conjugate gradient optimization algorithm using these gradient values and a cubic line search (Matlab Library, Carl E. Rasmussen, 2006).

### 3.3. Quasi-Newton Method: BFGS

Quasi-Newton methods (for example, BFGS, LBFGS, etc.) approximate the Hessian using gradient values and usually converge faster than first order gradient based methods. However, these methods (without line search) lack the convergence guarantee of EM and require line search that introduces additional computation. We implement BFGS using the same gradient functions as for conjugate gradient and the Matlab Optimization Toolbox, and compare with our algorithm.

## 4. Extension to Dirichlet Process Mixture Model

The Dirichlet process mixture model (DPMM) (Rasmussen, 2000) has gained significant attention for flexible density estimation. Unlike the traditional mixture models, DPMM does not assume a fixed number of density components and allows the model complexity to grow as more data points are seen. We have extended the variational Bayesian Dirichlet process mixture model (Blei & Jordan, 2006) using Deterministic Anti-annealing. The variational Bayesian DPMM is based on the truncated stick-breaking representation. Let $X$ represent the random variable that follows the Dirichlet process mixture model, and let the latent variables be $\mathbf{W} = \{\mathbf{V}, \boldsymbol{\eta}^*, \mathbf{Z}\}$ representing the stick lengths, individual mixture component parameters, and cluster memberships respectively. Let $q(\mathbf{v}, \boldsymbol{\eta}^*, \mathbf{z})$ denote the factorized variational distribution and $T$ be the truncation parameter. The posterior responsibility of a data point $\mathbf{x}_t$ to a mixture component $j \in \{1, \ldots, T\}$ is represented as $\phi_{ij}$, which is computed as:

$$\phi_{ij} = \frac{\exp(S_j)}{\sum_l \exp(S_l)} \quad (3)$$

where $S_j = E_q[\log V_j] + \sum_{l=1}^{j-1} E_q[\log(1 - V_l)] + E_q[\log p(\mathbf{x}_i | \mathbf{z}_i)]$.

The deterministic annealing approach can be easily extended to DPMM by a straight-forward modification of equation (3) (Katahira et al., 2008):

$$\phi_{ij} = \frac{\exp(S_j)^\beta}{\sum_l \exp(S_l)^\beta} \quad (4)$$

## 5. Results and Discussion

### 5.1. Datasets

We experiment with three different datasets.[1] The first dataset consists of samples drawn from a two component mixture of Gaussians, where the cluster sizes are 200k and 200 data points respectively (Figure 7(a)). We deliberately set the means and covariances so that there exists some overlap among the mixture components. The second dataset is also composed of a synthetic mixture of 4 Gaussians (Figure 7(b)), having 150k, 100k, 50k, and 150 data points each. Finally, we experiment with the MNIST handwritten digits dataset, which consists of high dimensional images of handwritten digits (0-9). We have randomly selected 5000 images of handwritten digit '4', and 250 images of handwritten digit '8'. Then we combine these samples and reduce the dimensionality of the combined dataset down to two dimensions using PCA. We choose these two particular digits because of their nice elliptical shape in 2D PCA projection (Figure 7(c)). We can approximate the density of this dataset by fitting a 2-component mixture of Gaussians.

### 5.2. Experimental Results

We observe the convergence behavior of all the four different algorithms: EM, Deterministic Anti-annealing, BFGS, and ECG, on all the three datasets described above. Each algorithm terminates when it satisfies the stopping criterion: $|L(\Theta^{k+1}) - L(\Theta^k)|/|L(\Theta^{k+1})| < \tau$. Here, $L(\Theta^k)$ represents the log-likelihood value at the $k^{th}$ iteration and $\tau$ is the tolerance parameter. For all the algorithms except Determinisitc Anti-annealing, we set the tolerance variable $\tau = 10^{-10}$. For Deterministic Anti-annealing, we set the tolerance parameter to $10^{-6}$. Since anti-annealing is capable of speeding up convergence at the later scheduling stages, we do not need a conservative tolerance. For the first and third experiment, the temperature schedule is set to $\beta = [0.8, 1.0, 1.2, 1.0]$, and for the second dataset we set $\beta = [0.2, 0.4, 0.6, 0.8, 1.0, 1.2, 1.0]$. As expected, we require slower temperature scheduling for a larger number of clusters. We estimate error with respect to the true parameter values by summing up the symmetric KL divergence between estimated and true Gaussian parameters for each mixture component. In order to measure the accuracy of clustering, we need to perform an assignment task to match clusters. For this task, we use the symmetric KL divergence between multivariate Gaussians and then perform a minimum weighted bipartite matching that finds a one-to-one mapping that minimizes the cumulative error with respect to true clustering. The symmetric KL divergence is defined as:

$$\mathcal{D}_s[p, q] = \mathcal{D}_{KL}(p, q) + \mathcal{D}_{KL}(q, p). \quad (5)$$

For two Gaussian distributions,

$$\mathcal{D}_s[\mathcal{N}(\mathbf{x}|\mu_i, \Sigma_i), \mathcal{N}(\mathbf{x}|\mu_j, \Sigma_j)] = \frac{1}{2}Tr\left[\Sigma_i^{-1}\Sigma_j + \Sigma_j^{-1}\Sigma_i\right]$$
$$+ \frac{1}{2}(\mu_i - \mu_j)^T\left[\Sigma_i^{-1} + \Sigma_j^{-1}\right](\mu_i - \mu_j) - d$$

The approximate error for the estimated parameter $\hat{\Theta} = \{\hat{\boldsymbol{\mu}}_j, \hat{\boldsymbol{\Sigma}}_j, \hat{\alpha}_j\}_{j=1}^K$ with respect to the true parameters $\Theta^*$ is estimated as follows:

$$err(\hat{\Theta}, \Theta^*) = \sum_{j=1}^K \mathcal{D}_s\left[\mathcal{N}(\mathbf{x}|\hat{\boldsymbol{\mu}}_j, \hat{\boldsymbol{\Sigma}}_j), \mathcal{N}(\mathbf{x}|\boldsymbol{\mu}^*_{\pi_j}, \boldsymbol{\Sigma}^*_{\pi_j})\right]$$
(6)

where $\{\pi_j\}_{j=1}^K$ is the one-to-one mapping estimated by minimum weight bipartite graph matching.

For each of the datasets, we apply all four algorithms for Gaussian mixture model fitting. We execute each algorithm 10 times on each dataset and observe the rate of convergence both for the best case (minimum error) and average case, as shown in Figure 7.[2] For EM and deterministic anti-annealing, we initialize the means of mixture components with randomly chosen sample points from the data. The mixing coefficients and covariances are initialized to the uniform distribution $(1/K)$ and the covariance of entire data respectively. For BFGS and ECG, we initialize the parameters with the outcome of a few EM iterations.

We experimented with the Dirichlet process Gaussian mixture models on synthetic Gaussian mixture models with high dynamic ranges in cluster sizes. We created a two-component Gaussian mixture, where the larger component consists of 50k points, and the smaller component consists of only 100 points. Then, we applied both the variational Bayes DPMM and the Deterministic Anti-annealing based DPMM. The VB-DPMM incorrectly estimated five components, where four components were fitted to the same Gaussian component. On the other hand, Anti-annealing DPMM resulted in the correct estimation of two components. The reason is the winner-take-all behavior during the anti-annealing.

---

[1] We also present additional results for image segmentation task in Appendix A.3.

[2] The plot showing the distribution of final log-likelihood values is presented in the Appendix A.1.

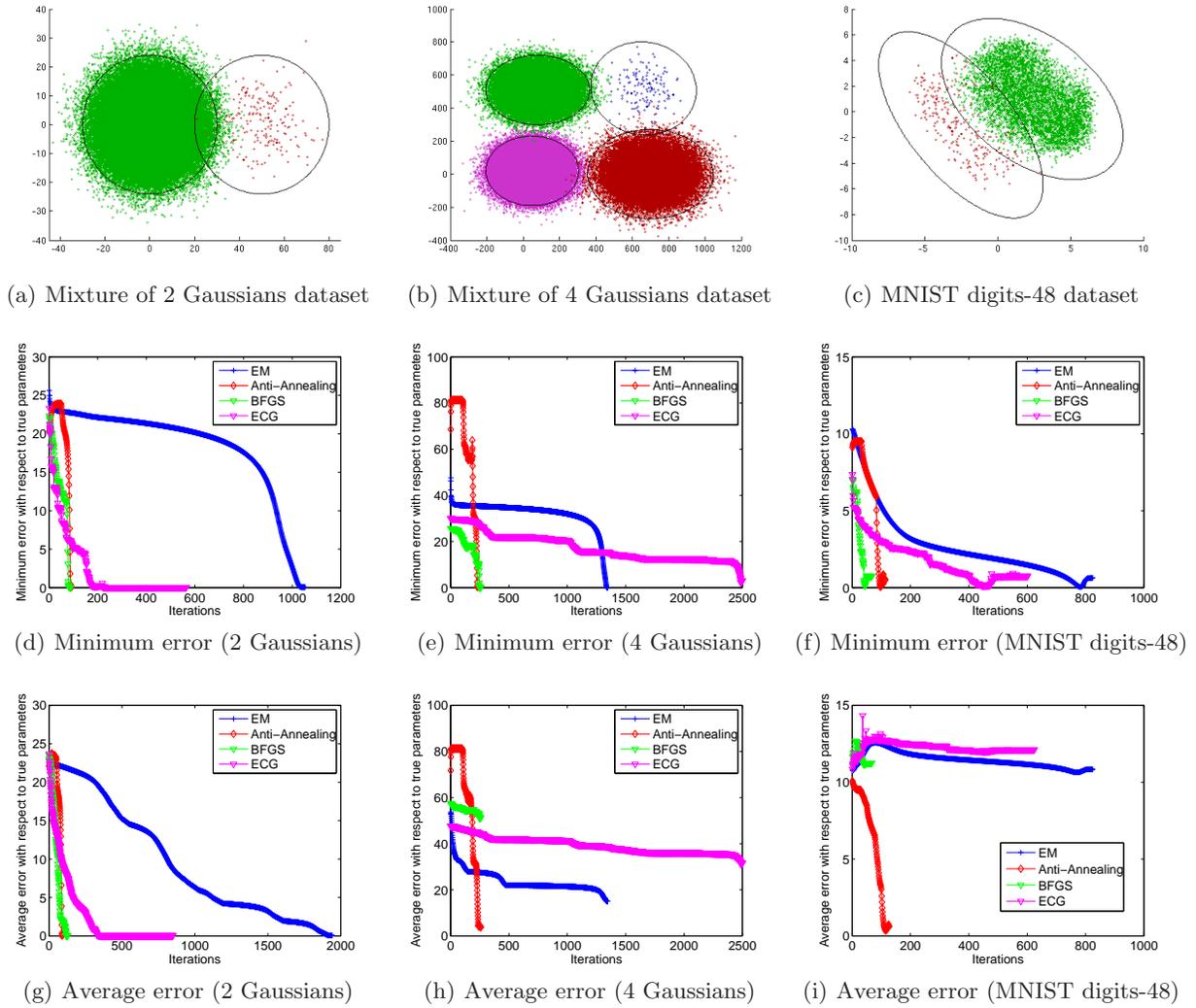

Figure 7. The speed of convergence of all the four algorithms in terms of error from true parameter values.

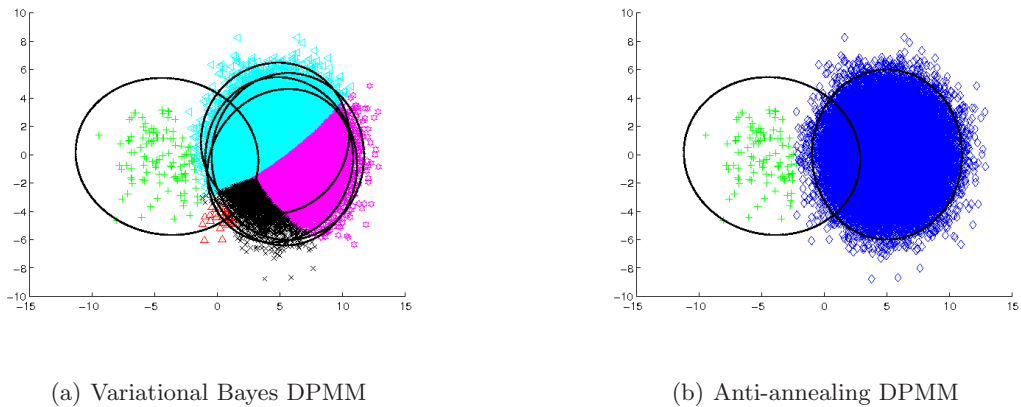

Figure 8. Comparison between traditional Variational Bayes DPMM against Deterministic Anti-annealing based DPMM.

Convergence of the EM Algorithm for Gaussian Mixtures with Unbalanced Mixing Coefficients### 5.3. Discussion

The experimental results demonstrate that both the deterministic anti-annealing method and BFGS significantly improve the speed of convergence compared to traditional EM. For the minimum error case scenario, BFGS outperforms deterministic anti-annealing by a small margin. However, deterministic anti-annealing is more stable on average and has better convergence behavior. Both BFGS and ECG exhibit significant amount of variability, particularly for the second (4 Gaussians) dataset. They often end up in poor degenerated local optima, where one or more of the mixing coefficients are clamped to zero. Therefore, deterministic anti-annealing outperforms BFGS on average. Salakhutdinov et al. (2003) proposed a hybrid EM-ECG algorithm, that estimates the entropy of cluster memberships as a measure of missing information (in other words, cluster overlap), and chooses to apply ECG if the entropy value is larger than certain threshold. Although the entropy-based method works well for balanced but highly overlapping mixtures, it is not general enough for the case of unbalanced mixtures. The entropy value decreases with the increasing skew in the mixing coefficients. Moreover, our experimental results show that neither ECG nor EM works well for such unbalanced overlapping mixtures.

## 6. Conclusion

The proposed Deterministic Anti-annealing scheme has lots of potential for faster convergence for datasets with smaller clusters. It offers several key advantages such as: 1) more robust global convergence, 2) faster convergence for small clusters via anti-annealing, 3) simple to implement, no line search required, 4) parameter constraints are satisfied without requiring reparameterization. Our experimental results demonstrate that deterministic anti-annealing EM outperforms all the other three algorithms on average, both in terms of speed and correctness of convergence. However, the temperature scheduling of Deterministic Annealing often requires some tuning (Rangarajan, 2000). A thorough study of temperature schedule can be an interesting future direction. As a general guideline, the schedule should be guided by the complexity of the data. The more complex the data, the more slowly we should vary the temperature parameter.

## 7. Acknowledgement

We would like to thank Daniel Štefankovič, Gaurav Sharma, and Suprakash Datta for many useful comments and feedback. Funded in part by NSF award IIS-0910611.

## References

Blei, D.M. and Jordan, M.I. Variational inference for dirichlet process mixtures. *Bayesian Analysis*, 1(1):121–144, 2006.

Dempster, A.P., Laird, N.M., and Rubin, D.B. Maximum likelihood from incomplete data via the EM algorithm. *Journal of the Royal Statistical Society. Series B (Methodological)*, 39(1):1–38, 1977.

Jain, A.K., Murty, M.N., and Flynn, P.J. Data clustering: a review. *ACM computing surveys (CSUR)*, 31(3):264–323, 1999.

Katahira, K., Watanabe, K., and Okada, M. Deterministic annealing variant of variational bayes method. In *Journal of Physics: Conference Series*, volume 95, pp. 012015. IOP Publishing, 2008.

Kearns, M.J., Mansour, Y., and Ng, A.Y. An information-theoretic analysis of hard and soft assignment methods for clustering. *Nato ASI Series, Series D: Behavioural and Social Sciences*, 89:495–520, 1998.

Liu, D.C. and Nocedal, J. On the limited memory BFGS method for large scale optimization. *Mathematical programming*, 45(1):503–528, 1989.

Ma, J., Xu, L., and Jordan, M.I. Asymptotic convergence rate of the EM algorithm for Gaussian mixtures. *Neural Computation*, 12(12):2881–2907, 2000.

McLachlan, G. and Peel, D. *Finite Mixture Models*. New York: Wiley Interscience, 2000.

Rangarajan, A. Self-annealing and self-annihilation: unifying deterministic annealing and relaxation labeling. *Pattern Recognition*, 33(4):635–649, 2000. ISSN 0031-3203.

Rasmussen, C.E. The infinite Gaussian mixture model. *Advances in neural information processing systems*, 12:554–560, 2000.

Redner, R A and Walker, H F. Mixture densities, maximum likelihood and the EM algorithm. *SIAM Review*, 26(2):195–239, 1984.

Rose, K. Deterministic annealing for clustering, compression, classification, regression, and related optimization problems. *Proceedings of the IEEE*, 86(11):2210–2239, 1998. ISSN 0018-9219.

Salakhutdinov, R., Roweis, S., and Ghahramani, Z. Optimization with EM and Expectation-Conjugate-Gradient. *in Proceedings of ICML*, 20:672–679, 2003.

Ueda, N. and Nakano, R. Deterministic annealing EM algorithm. *Neural Networks*, 11(2):271–282, 1998.

Wu, CF. On the convergence properties of the EM algorithm. *The Annals of Statistics*, 11(1):95–103, 1983.

Xu, L. and Jordan, M.I. On convergence properties of the EM algorithm for Gaussian mixtures. *Neural computation*, 8(1):129–151, 1996.